\documentclass{article}
\pdfoutput=1

\usepackage{arxiv}
\usepackage[utf8]{inputenc} 
\usepackage[T1]{fontenc}    
\usepackage{float}
\usepackage{hyperref}       
\usepackage{url}            
\usepackage{booktabs}       
\usepackage{amsfonts}       
\usepackage{nicefrac}       
\usepackage{microtype}      
\usepackage{lipsum}		
\usepackage{graphicx}
\usepackage{doi}

\usepackage{wrapfig}        
\usepackage{graphicx}
\usepackage{subfigure}
\usepackage{authblk}
\newcommand*{\affaddr}[1]{#1} 

\newcommand*{\email}[1]{\texttt{#1}}

\title{ViF-SD2E: A Robust Weakly-Supervised Method for Neural Decoding}

\date{} 					

\author{%
\textbf{
Jingyi Feng, Yong Luo, and Shuang Song
}\\
\affaddr{Institute of Artificial Intelligence, School of Computer Science, Wuhan University}\\
\email{\{fjy2035, luoyong, songshuang327\}@whu.edu.cn}\\
}

\renewcommand{\headeright}{}
\renewcommand{\undertitle}{\large{Subtitle: From Robust Vision-feedback to Brain}}
\renewcommand{\shorttitle}{ViF-SD2E: A Robust Weakly-Supervised Method for Neural Decoding}

\hypersetup{
pdftitle={A template for the arxiv style},
pdfsubject={q-bio.NC, q-bio.QM},
pdfauthor={David S.~Hippocampus, Elias D.~Striatum},
pdfkeywords={First keyword, Second keyword, More},
}

\begin{document}
\maketitle

\begin{abstract}
\label{abstract}
Neural decoding plays a vital role in the interaction between the brain and the outside world.
In this paper, we directly decode the movement track of a finger based on the neural signals of a macaque.
Supervised regression methods may overfit to actual labels containing noise, and require a high labeling cost, while unsupervised approaches often have unsatisfactory accuracy.
Besides, the spatial and temporal information is often ignored or not well exploited by those methods.
This motivates us to propose a robust weakly-supervised method, called ViF-SD2E, for neural decoding.
In particular, it consists of a space-division (SD) module and a exploration--exploitation (2E) strategy, to effectively exploit both the spatial information of the outside world and the temporal information of neural activity, where the SD2E output is analogized with the weak 0/1 vision-feedback (ViF) label for training.
It is worth noting that the designed ViF-SD2E is based on a symmetric phenomenon between the unsupervised decoding trajectory and the real trajectory in previous observations, then a cognitive pattern of fuzzy (robust) interaction in the nervous system may be discovered by us.
Extensive experiments demonstrate the effectiveness of our method, which can be sometimes comparable to supervised counterparts.
\end{abstract}

\keywords{Neural decoding, weakly-supervised, space division, vision-feedback}

\section{Introduction}
\label{introduction}

Neural coding and decoding are vital technologies for realizing the brain--computer interface, with diverse potential applications, e.g., facilitating the daily life of paralyzed patients \cite{wallisch2014matlab}.
Existing studies mainly focus on movement, speech, and vision, aiming to gain a scientific understanding of the link between neural activity and the outside world \cite{livezey2021deep}.
This paper focuses on locating a macaque's moving finger by decoding the neural spike signals.
Addressing this problem not only has direct positive impacts on the impaired user's ability to communicate with the world, but also provides a solution for human--computer interaction for healthy users \cite{fan2018thoughts}.
Furthermore, prostheses, robots, mice, and other devices that fully realize `brain control technology' are becoming reality \cite{mcfarland2008brain, kim2016commanding}.
For people with a physical disability, a prosthesis can be installed and controlled through neural decoding \cite{hamedi2016electroencephalographic}.

The problem of decoding finger movement is described in a pioneer work \cite{georgopoulos1989mental}, which finds that there is a correspondence between the direction and location of the movement of the upper limb of a macaque and the spike signal in its motor cortex.
Most of the early work on neural decoding are time-independent \cite{xue2017unsupervised, paranjape2019cross}.
However, since the movement is a continuous process, more recent work focuses on exploring time correlations \cite{ali2017robust, wu2019neural}.
In recent years, neural network approaches, such as recurrent neural networks and long short-term memory \cite{xie2018decoding, ahmadi2019decoding}, have achieved remarkable performance for neural decoding.
However, most of these approaches are often supervised, which may lead to over-fitting to noisy target values (e.g., finger positions).
Although there are also a few semi-supervised \cite{bishop2013semi}, unsupervised \cite{xue2017unsupervised}, and weakly supervised decoding \cite{feng2020weakly} approaches, the spatial and temporal information connecting the external world to neural activity are ignored or not well exploited.

\begin{figure}    
    \centering
    \subfigure[]{
        \includegraphics[width=0.41\textwidth]{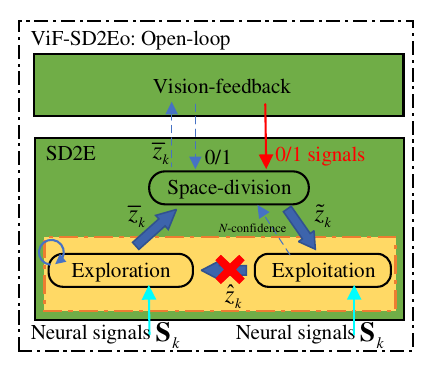}
    }
    \subfigure[]{
        \includegraphics[width=0.41\textwidth]{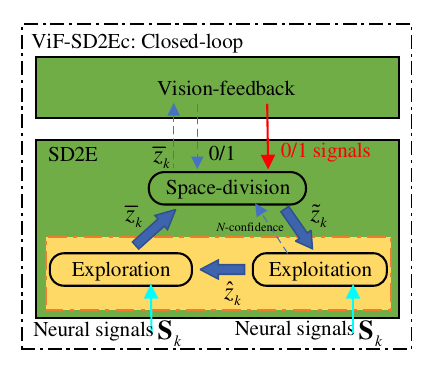}
    }
    \caption{Open-loop (a) and closed-loop (b) optimization procedures of the proposed ViF-SD2E method. Here, the red arrow and the cyan arrow represent the input of vision-feedback from the outside world and the input of neural signals from the brain, respectively. The solid arrow represents the data transmission process and the dashed arrow represents interactive and evaluative feedback, which constitute the training or learning process of the ViF-SD2E.
    $\textbf S_k$ is the neural signal after time series processing. $z_k$ is the true label. $\hat{z}_k$ is the value decoded by unsupervised exploration. $\bar{z}_k$ is the value corrected by SD module. $\tilde{z}_k$ is the value output by supervised exploitation and is used to update the weights of exploration. $k$ is the $k$th moment or the $k$th sample. The bits (0/1) are the converted symbols after being decoded by SD2E (neural signals $\rightarrow$ value decoded $\rightarrow$ 0/1), and then analogized with the symbols (0 or 1) provided from ViF.
    }
    \label{fig:ViF-SD2E_Procedure}
\end{figure}

To remedy these drawbacks, we propose a novel weakly-supervised neural decoding method, called ViF-SD2E, which mainly consists of a spatial division (SD) module together with an unsupervised exploration and supervised exploitation (2E) strategy, where only weak 0/1 vision-feedback (ViF) labels are required for training.
In particular, the input neural signals are sent to an unsupervised expectation maximization (EM) model to \textbf{explore} (induce) some initial prediction results.
Since the obtained results are often unreliable, we divide the observable space into several non-overlapping regions, and assume that there are some weakly-supervised 0/1 labels (vision feedback of the macaque), which indicate the regions in which the target values are located.
Here, the space can be divided at different resolutions depending on the degree of supervision.
By comparing the binarized initial results and the weak labels, we obtain some corrected outputs.
These outputs are then 
\textbf{exploited} in a supervised manner to refine the prediction results, where the temporal information is also exploited.
The refined results can be used directly as the final prediction results (open-loop), or to update the parameters of the EM model.
The latter leads to a closed-loop optimization, and the exploitation and exploration should iterate until convergence.
Figure \ref{fig:ViF-SD2E_Procedure} is an illustration of the open-loop and closed-loop optimization procedures of the proposed ViF-SD2E method. 
Its specific implementation is concerning Figure \ref{fig:ViF-SD2E_diagram} and Figure \ref{fig:ViF-SD2E_gl}.
The advantages and disadvantages between open-loop and closed-loop are analyzed concerning Chapter \ref{computational complexity}. 
What's more, (a) and (b) are two simplified schemes designed by us to simulate the neural decoding process, based on a factual finding \cite{feng2020weakly}.

The main contributions of this paper are summarized as follows.
\begin{itemize}
\item We propose a robust weakly-supervised method for movement decoding via neural signals.
Only robust 0/1 signals from the ViF are required for model training.
\item We design a novel SD module together with a 2E strategy to effectively make use of both the spatial and temporal information for the neural decoding, where the degree of supervision can be flexibly controlled.
\item Both open-loop and closed-loop algorithms are developed for optimization.
\end{itemize}

We conducted extensive experiments on a popular neural decoding dataset.
The results demonstrate that our method is superior to other competitive approaches in most cases, and can be comparable or even outperform some supervised counterparts.

\section{Related work}

In our point of view,  “visual feedback” can also include any perceptions or external stimulus in the embodied mind, such as from the eyes, ears, nose, tongue and body.
It provides visual feedback by receiving information from the outside world.
In this paper, we focus on visual object detection, because it is important to locate the target in the motion.
In the beginning, many technologies were biased towards supervised methods, then came target detection that applied reinforcement learning \cite{zhang2020from}, and later, unsupervised and self-supervised technology \cite{2021DetCo}.
Recently, interactive target detection technology has been a significant innovation \cite{chen2021pix2seq}.
Last but not least, in our opinion, the working mechanism of the brain's cognition should include interaction and reinforcement. Here, “reinforcement” refers to the independent thinking of the brain using existing memory or experience (Exploitation), which can also be said to be self-reinforcement.

In addition, state-space models (SSMs) with associations between the current state and the previous state became more developed \cite{gilja2012a, wallisch2014matlab} in neural decoding.
Shanechi et al. gave an SSM model and used optimal feedback control to decode the movement state \cite{shanechi2013feedback}.
Based on that SSM model, Wu et al. derived a convolutional space model (CSM), which correlates the current state with the neural signals at multiple previous moments \cite{wu2019neural}.
In recent years, deep learning has been the object of a large number of studies in neural decoding, such as RNN \cite{tseng2019decoding} and LSTM \cite{elango2016sequence, pan2018rapid}.
The decoding performed by these methods is more accurate than that of the independent linear method.

Furthermore, Sussillo et al. found that the source of the recorded neural activity can change from  day to day, e.g., due to a slight movement of the implanted electrodes.
The proposed multiplicative RNN allows mappings from the neural input to the motor output to partially change from neural activity \cite{sussillo2018making}.
Wu et al. proposed an unsupervised cubature Kalman filter (UCKF) by exploring the relations between the neural signals and the movement \cite{xue2017unsupervised}.
Due to the instability of this unsupervised model, Feng et al. found that the part/all of the unsupervised decoding positions was reversed in the movement region, leading to significant errors.
Then, they introduced a priori binary to verify the existence of this reverse mechanism \cite{feng2018neural, feng2020weakly}. However, this study is only a factual finding, and there is no effective and intuitive model structure and basis.

Therefore, we focus on the fact that the symmetric model was found in the paper [15], and that neural signals are temporal.
We propose the designed ViF-SD2E, and attempt to explain the rationality of the model from other scientific perspectives (neuroscience, machine learning, etc.) and further experimentally verify the validity of the model.
Its robustness is reflected in the application of symmetric mode, which has better fault tolerance for finger movement.

\section{The space-division and exploration--exploitation with vision-feedback (ViF-SD2E)}

\begin{figure*}
\centering
\includegraphics[width=0.80\textwidth]{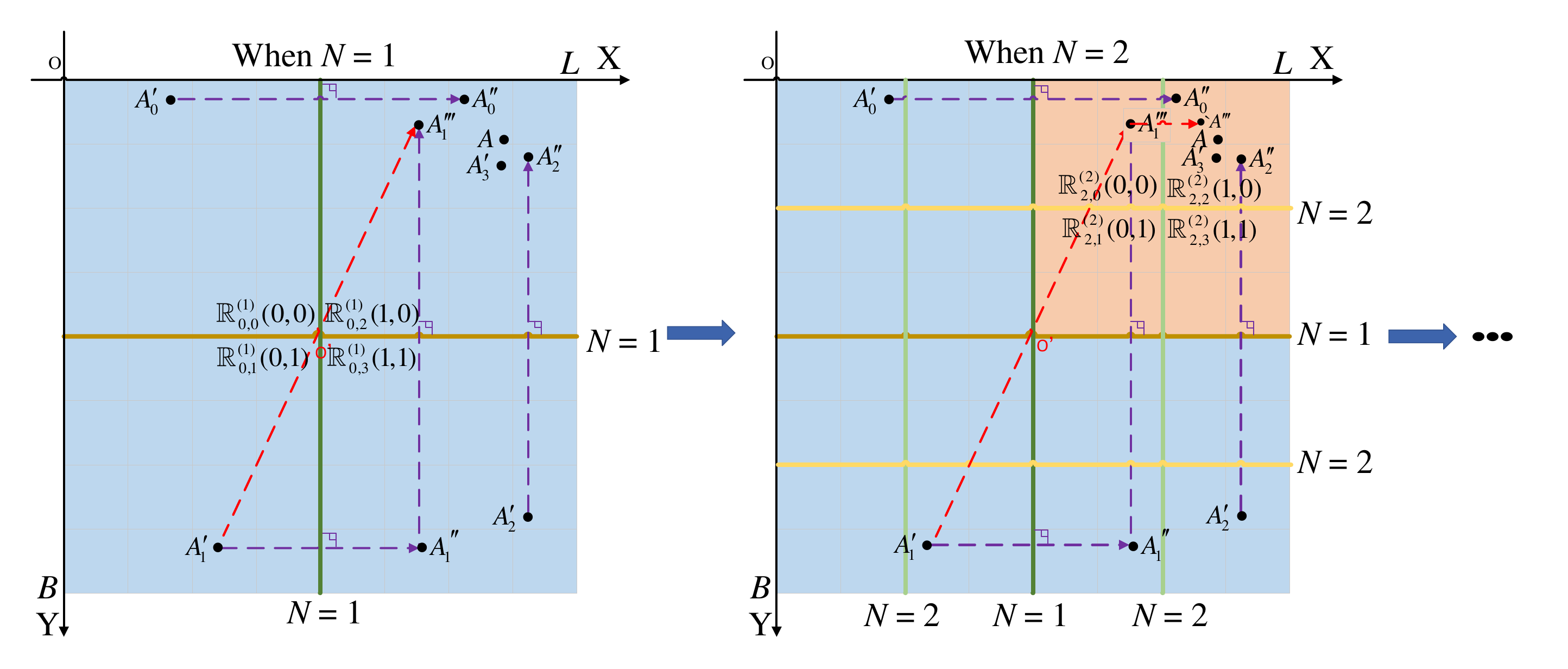}
\caption{The movement space of the finger that interacts with the brain.
Assume that the blue ($N=1$) and orange ($N=2$) area are the active space in a certain decoding.
The $A\bullet$ is the real label, and $A'_0\bullet$, $A'_1\bullet$, $A'_2\bullet$ or $A'_3\bullet$ are the unsupervised predicted values that may appear after the exploration.
Step 1 ($N=1$): The 2-D space is divided equally, and their intersection is $O'$.
Then four subspaces, $\mathbb R^{(1)}_{0,0}(0, 0)$, $\mathbb R^{(1)}_{0,1}(0, 1)$, $\mathbb R^{(1)}_{0,2}(1, 0)$ and $\mathbb R^{(1)}_{0,3}(1, 1)$, are demarcated.
Next, we use spatial symmetry to keep the predicted values and the true labels in one subspace, and obtain $\bullet A''_0$, $\bullet A'''_1$, $\bullet A''_2$ and $\bullet A'_3$.
Step 2 ($N=2$): Repeat step 1: $\mathbb R^{(1)}_{0,2}(1, 0)$ is divided into four subspaces, $\mathbb R^{(2)}_{2,0}(0, 0)$, $\mathbb R^{(2)}_{2,1}(0, 1)$, $\mathbb R^{(2)}_{2,2}(1, 0)$, $\mathbb R^{(2)}_{2,3}(1, 1)$.
Here, $A\bullet$ is located in $\mathbb R^{(2)}_{2,2}(1, 0)$.
Then we also employ spatial symmetry to keep them in one subspace, and get $\bullet A''_0$, $\bullet A''''_1$, $\bullet A''_2$ and $\bullet A'_3$.
As $N\rightarrow + \infty$, our predicted values will gradually approach the true labels.}
\label{fig:ViF-SD2E_movementspace-1}
\end{figure*}

\begin{figure*}
\centering
\includegraphics[width=0.80\textwidth]{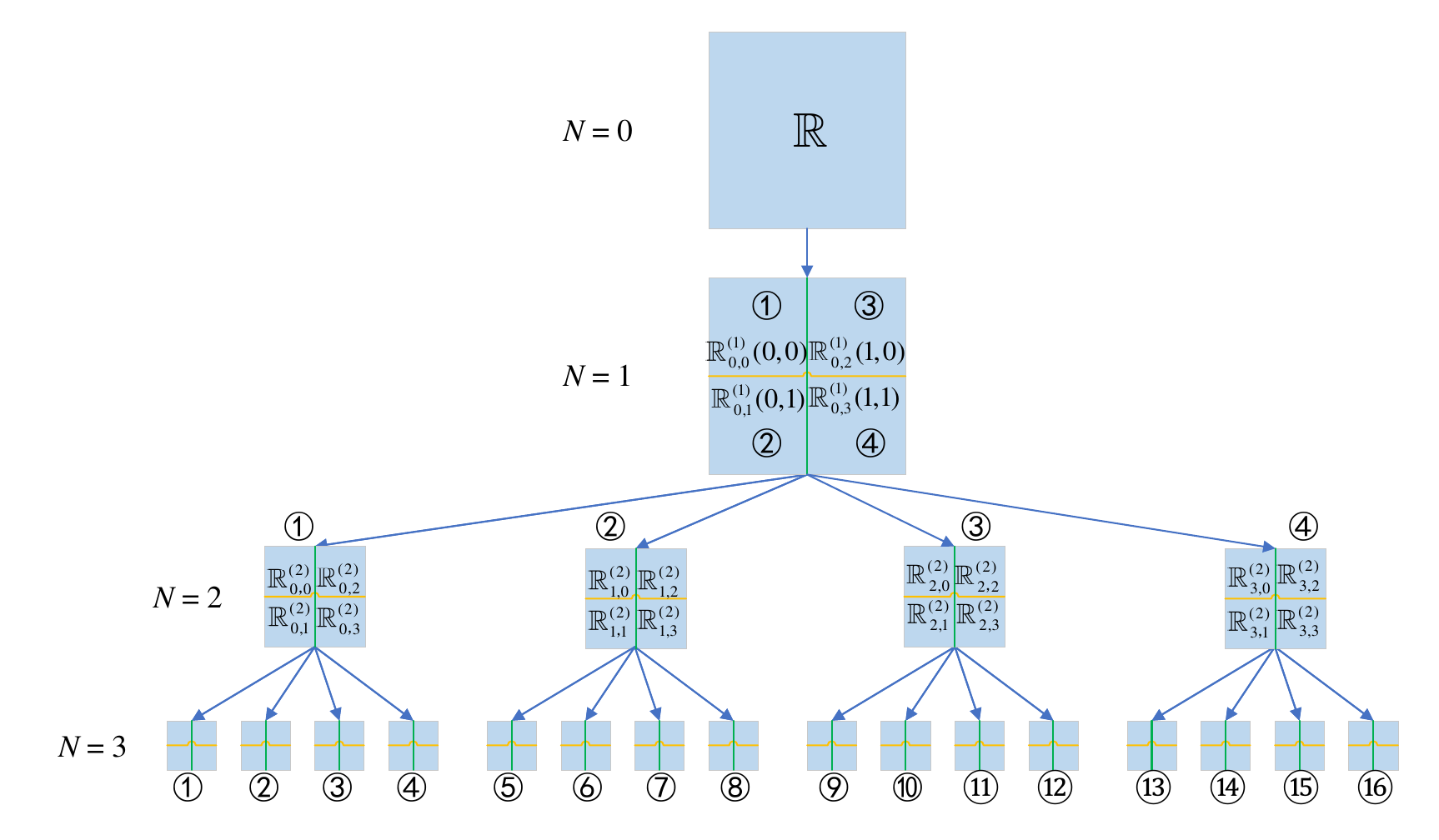}
\caption{Intuitive diagram of space division in the movement space.}
\label{fig:ViF-SD2E_movementspace-2}
\end{figure*}

The proposed ViF-SD2E is mainly derived from an experimental finding: that is, the unsupervised predicted values in the exploration and the actual labels have partial or full symmetry in the 2-D space \cite{feng2020weakly}.
The function of the vision-feedback (ViF) is to provide effective 0/1 feedback to the SD2E; the function of the SD module is to process the information from the exploration and to be a tool for a state transition area from the ViF to the SD2E; the function of the exploration from the 2E strategy is to be a tool for unsupervised exploration; the function of the exploitation from the 2E strategy is to be a tool for supervised training.
Finally, trained exploitation is used for testing and further used to evaluate the $N$-confidence of the corrected position.

\subsection{The vision-feedback (ViF) from the outside world and coding mechanism}

First, the movement space has length $L$ and width $B$, such as the size of the screen or that of the receptive field.
The origin is the upper left corner.
Like vision-feedback from humans, we can know which subspace the finger has moved, but its location cannot be readily given by us in a robust fashion.
In this paper, the actual labels are encoded as 0/1 in the movement space/subspace and used to extract the maximum and minimum values of the active space, which are used as the ViF.
So the total number of subspace is $S(N)=2^{2N}$.
In addition, the maximum fault tolerance of the ViF-SD2E is $R(N)=L/2^N$ along the $x$-axis and $R(N)=B/2^N$ along the $y$-axis (see subsection \ref{Correction and robustness}).
Here, $N$ is a spatial parameter to control the change from space to subspace.
This is shown in Figures \ref{fig:ViF-SD2E_movementspace-1} and \ref{fig:ViF-SD2E_movementspace-2}.

We will now explain the coding mechanism using an example.
The movement space is represented as $\mathbb R$.
The meaning of its subspace $\mathbb R^{(N)}_{n, i}(x_{bit}, y_{bit})$ is the $i$th quadrant of the $n$th subspace of the $N$th divided space,
where $N$ indicates the $N$th division of the space, $n$ is the $n$th subspace, $i$ is the $i$th quadrant, and $x_{bit}$ and $y_{bit}$ are the encoded $x$-position and $y$-position in the $xy$-plane,
as shown in Figure \ref{fig:ViF-SD2E_movementspace-2}.
When $N=0$, it is itself, that is, $\mathbb R$: $\mathbb R^{(0)}(0/1, 0/1)$.
When $N=1$, $\mathbb R$ has been equally divided into four subspaces, namely, $\mathbb R^{(1)}$: $\mathbb R^{(1)}_{0, 0}(0, 0)$, $\mathbb R^{(1)}_{0, 1}(0, 1)$, $\mathbb R^{(1)}_{0, 2}(1, 0)$, $\mathbb R^{(1)}_{0, 3}(1, 1)$.
When $N=2$, each subspace of $\mathbb R^{(1)}$ has been equally divided into four subspaces, and there are now 16 subspaces in all.
When $N=3$, $\mathbb R^{(2)}$ has been divided into 32 subspaces in all, and so on.
That is, if an actual $z_k$($x_k$, $y_k$) is encoded as\\
$z_k$: $\mathbb R^{(0)}(0/1, 0/1)$ $\rightarrow$ $\mathbb R^{(1)}_{0, 1}(0, 1)$ $\rightarrow$ $\mathbb R^{(2)}_{1, 3}(1, 1)$ $\rightarrow$ $\mathbb R^{(3)}_{4, 1}(0, 1)$ $\rightarrow$ $\cdots$,
then the $x$- and $y$-axes in the actual coordinate $z_k$ are encoded as
$x_k$: 0/1 $\rightarrow$ 0 $\rightarrow$ 1 $\rightarrow$ 0 $\rightarrow$ $\cdots$; $y_k$: 0/1 $\rightarrow$ 1 $\rightarrow$ 1 $\rightarrow$ 1 $\rightarrow$ $\cdots$.\\

Next, as shown above, the coding mechanism and symmetric correction are used to keep the predicted values and the true labels in the same subspace.
Therefore, the calculation of the movement encoder and corrector is given as follows:
\begin{equation}
    {\lim\limits_{{\mathbb R}: N\rightarrow +\infty}} F^{(N)}_{bit}=\left\{
             \begin{array}{llr}
                1  \quad if \; \overline z_k \;or\; z_k \geq f_{mid} \\
                0  \quad if \; \overline z_k \;or\; z_k  < f_{mid}
             \end{array}
    \right.
\end{equation}
\begin{equation}
    {\lim\limits_{{\mathbb R}: N\rightarrow +\infty}} F^{(N)}_{update}=\left\{
             \begin{array}{llr}
                \overline z_{k}  & if \; \overline z_{k,bit}=z_{k,bit}  \\
                2(f_{mid}-\overline z_{k})+\overline z_{k} & if \; \overline z_{k,bit} \neq z_{k,bit}
             \end{array}
    \right.
\end{equation}
here, ${\lim_{{\mathbb R}: N\rightarrow +\infty}} F^{(N)}_{bit}$ indicates that as the $N$-value is increased in $\mathbb R$, the position that is encoded approaches the subspace where the true label is located.
$F^{(N)}_{bit}$ are the encoded 0/1 signals, such as $\bar z_{k, bit}$ or $z_{k, bit}$ during the $N$th division.
$f_{mid}=(z_{max}+z_{min})/2$ is the central axis of the space/subspace, and $z_{max}$ and $z_{min}$ are the maximum and minimum boundaries of the movement during the $N$th division.
In the corrector, ${\lim_{{\mathbb R}: N\rightarrow +\infty}} F^{(N)}_{update}$ indicates that as the $N$-value is increased in $\mathbb R$, the position that is corrected approaches the true label.
$F^{(N)}_{update}$ is the corrected/updated value in the $N$th division.
Beside, we define the function $Z(\bullet)$, which represents the synergy of $F^{(N)}_{bit}$ and $F^{(N)}_{update}$.
$k$ refers to the $k$-th moment or the $k$-th sample.

As $N \rightarrow +\infty$, ViF-SD2E can be equivalent to a supervised model, and when $N=0$, it is an unsupervised mode.
When $N$ is set between these two, this can be understood as a weakly-supervised mode.
Here, formula (1) is used as an auxiliary of formula (2); the core of formula (2) is that there is a symmetry between $\overline z_k$ and the $z_k$ in the movement space/subspace.

\subsection{Spatio-temporal mining in ViF-SD2E}
\label{ViF-SD2E-overview}

\begin{figure}
\centering
\includegraphics[width=1.02\textwidth]{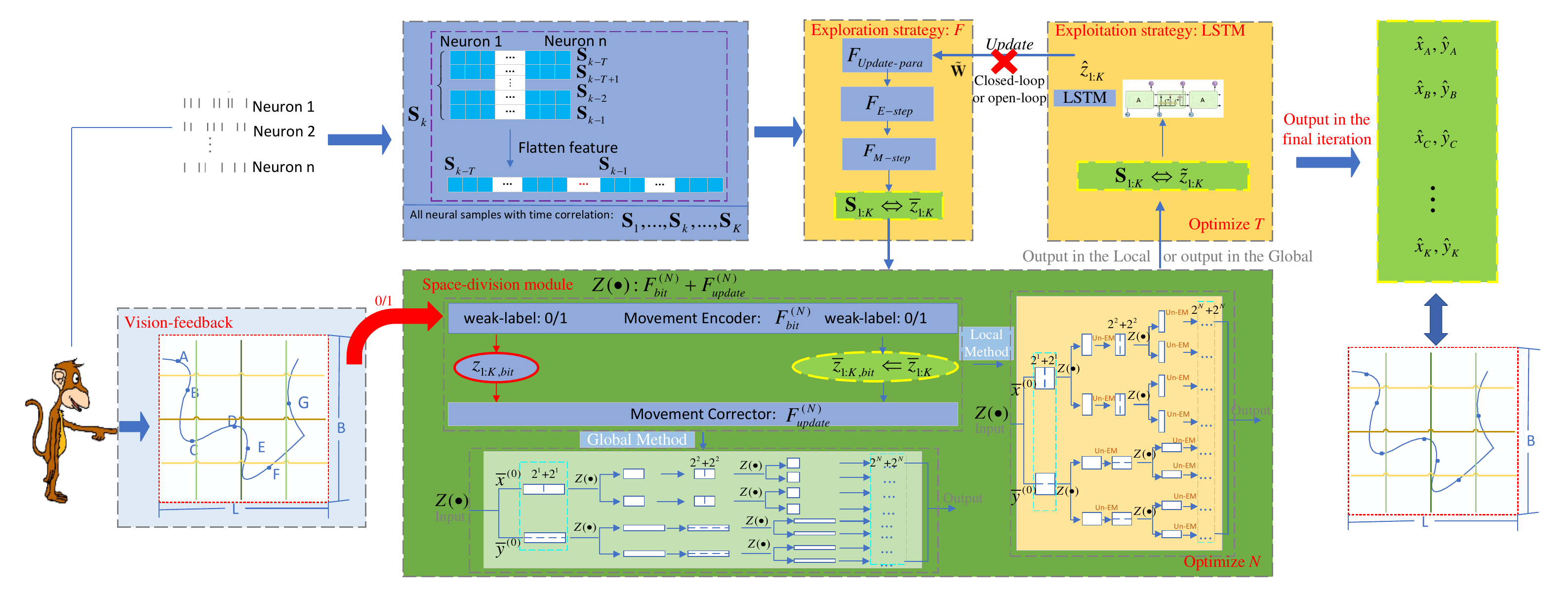}  
\caption{The framework of the proposed ViF-SD2E.
In the training phase: 1) the neural signals that have a one-to-one correspondence with the movement of the finger are collected, and the movements are encoded as 0/1 signals via the ViF; the neural signals corresponding to each sample are preprocessed as feature $\textbf S_{k}$ together with a time sequence, which is the input to our ViF-SD2E.
2) Then, the $\overline z_{1:K}$ predicted by the exploration are fed into the designed SD module, where they are encoded as $\overline z_{1:K, bit}$ and compared with the given $z_{1:K, bit}$.
After being processed by global or local methods, the $\overline z_{1:K}$ is corrected to $\tilde z_{1:K}$; 3) in the exploitation; the $\tilde z_{1:K}$ are then used as ground truth to train the exploitation together with the input feature ${\bf S}_{1:K}$.
Finally, the trained exploitation is adopted for testing and further used to evaluate the $N$-confidence.}
\label{fig:ViF-SD2E_diagram}
\end{figure}
\begin{figure}    
    \centering
    \subfigure[Global method]{
        \includegraphics[width=0.48\textwidth]{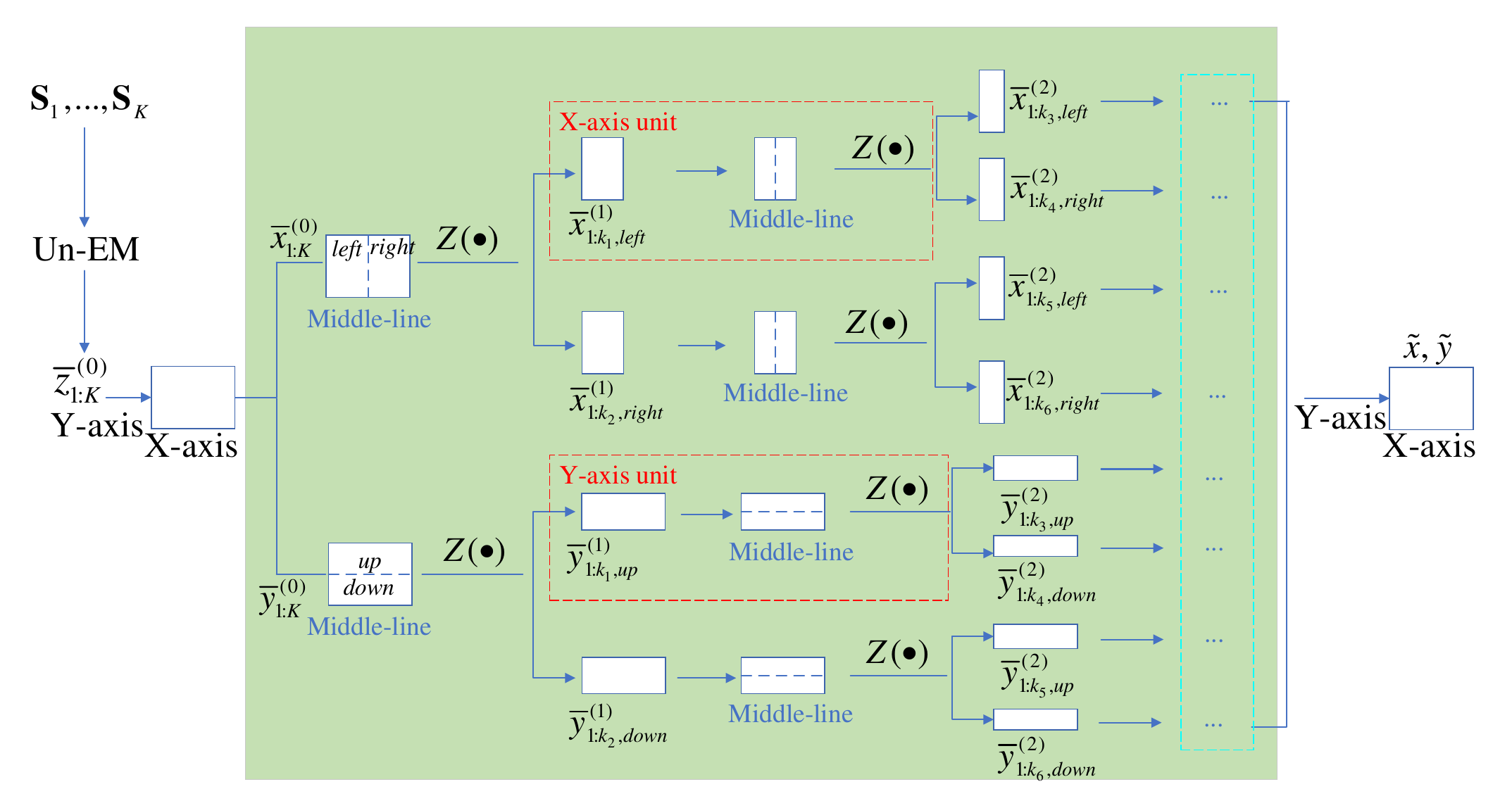}
    }
    \subfigure[Global unit]{
        \includegraphics[width=0.48\textwidth]{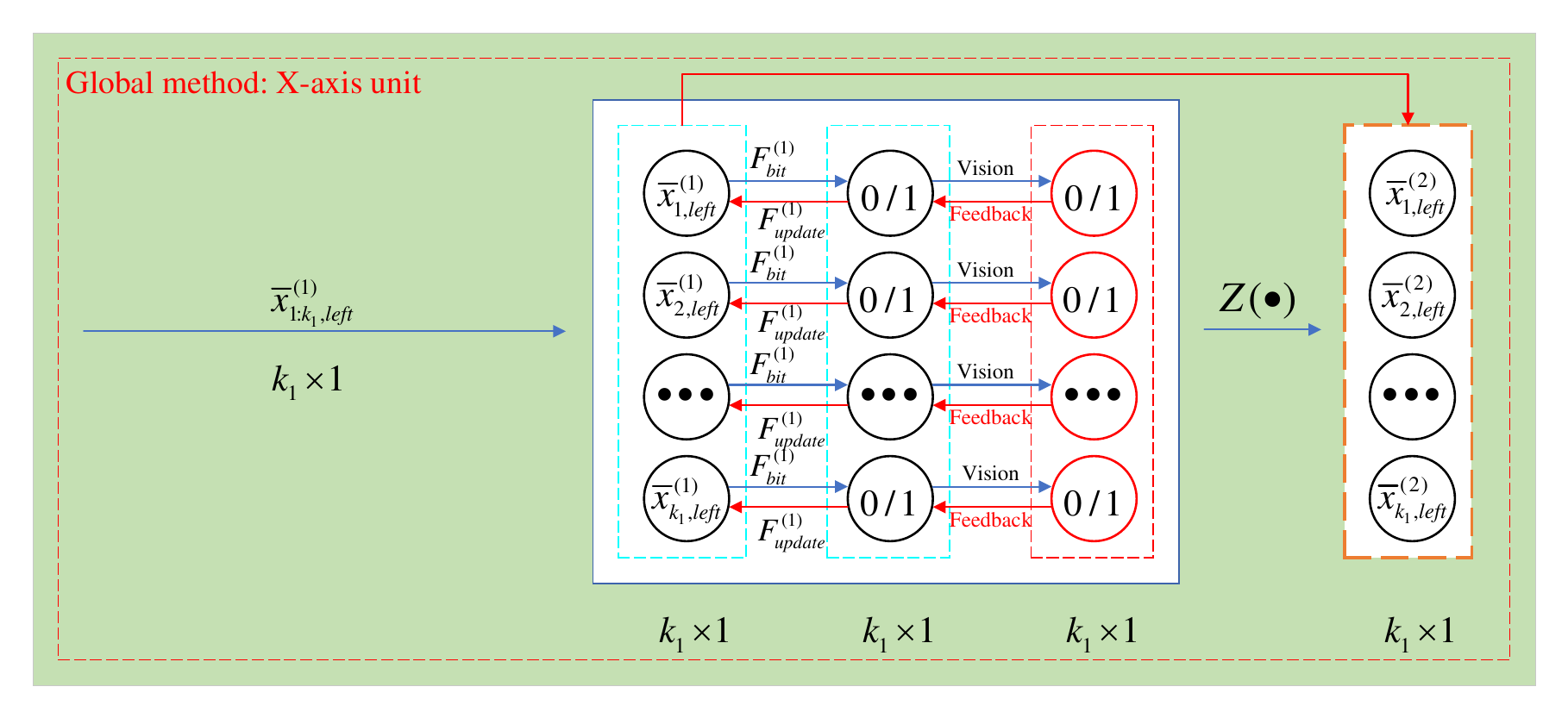}
    }
    \subfigure[Local method]{
        \includegraphics[width=0.48\textwidth]{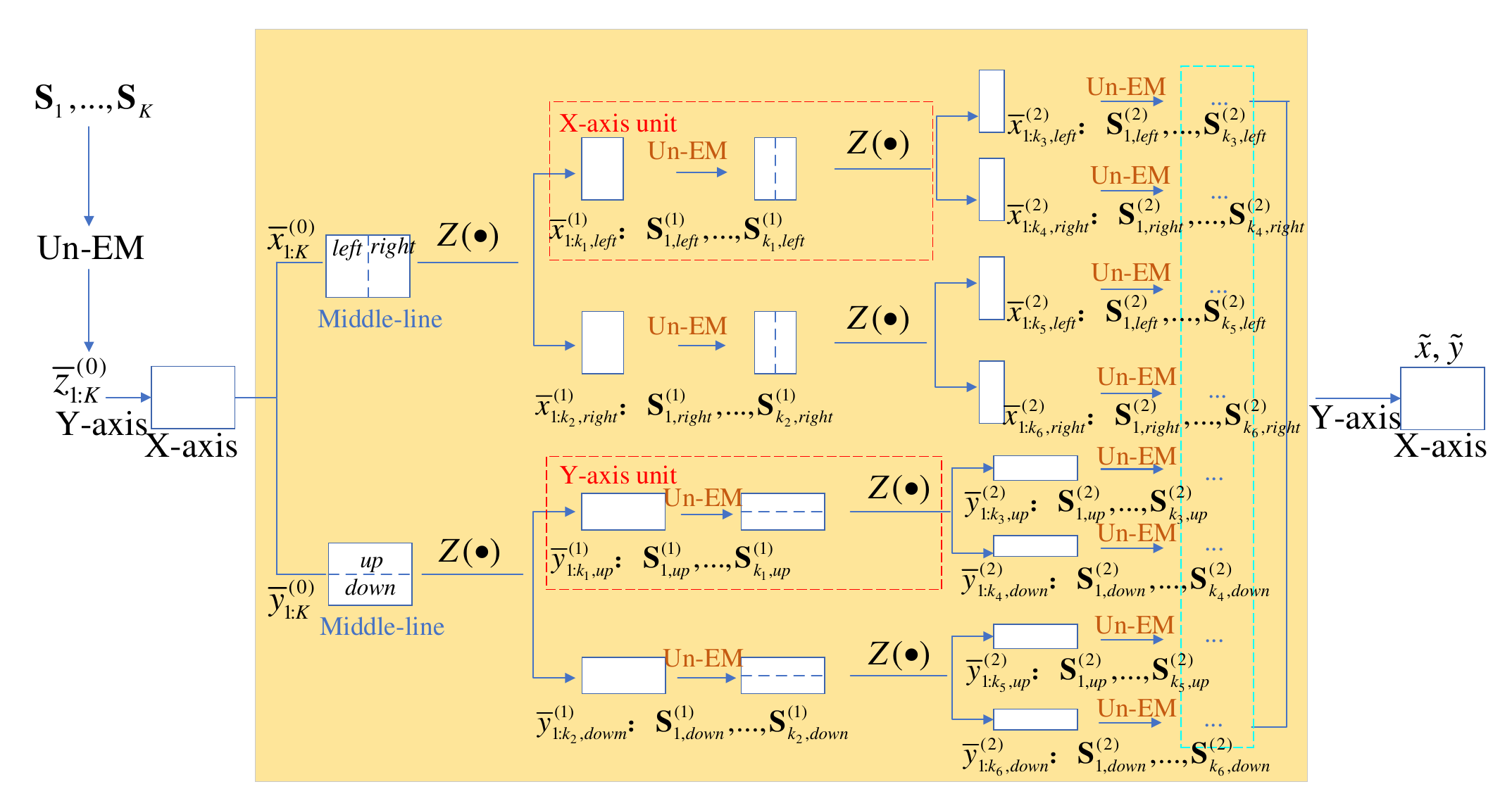}
    }
    \subfigure[Local unit]{
        \includegraphics[width=0.48\textwidth]{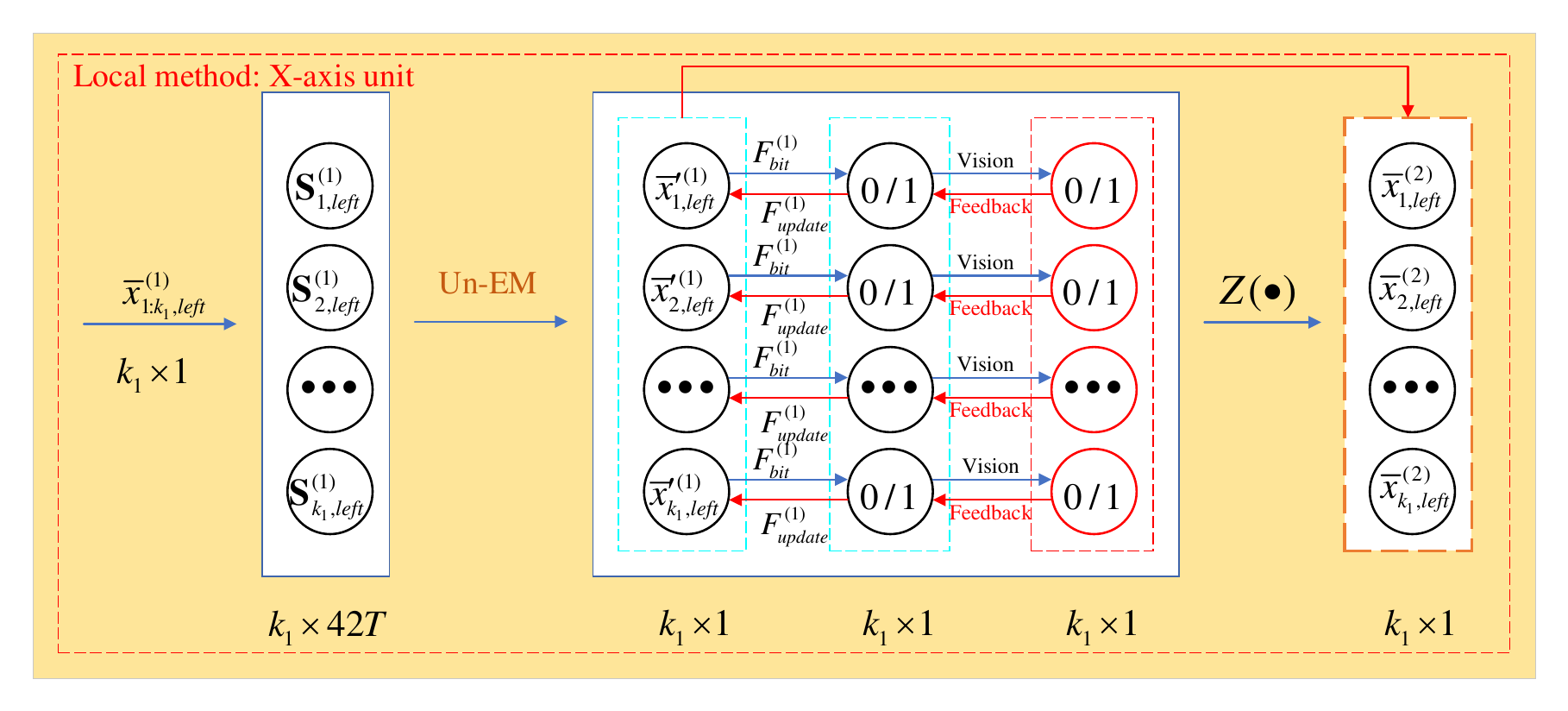}
    }
    \label{fig:ViF-SD2E_local}
    \caption{The red box represents the smallest processing unit in the global and local methods.
(a) The steps of the global method.
What the global method is concerned about is that the unsupervised predicted $\overline x^{(0)}_{1:K}$ and $\overline y^{(0)}_{1:K}$ are sent directly to the equation (1)-(2) for coding and correction.
Furthermore, the output of the global method is the corrected $\tilde x^{N}_{1:K}$ and $\tilde y^{N}_{1:K}$.
(b) The processing unit of the global method.
$\overline x^{(1)}_{1:k_1, left}$ indicates that in the first division, the samples of the $x$-axis in the left subspace are taken, and the total number of samples is $k_1$.
Next, $\overline x^{(1)}_{1:k_1, left}$ is coded as 0/1 by $F^{(1)}_{bit}$ and compared with the given 0/1, and then corrected by $F^{(1)}_{update}$.
Finally, the output is $\overline x^{(2)}_{1:k_1, left}$.
(c) The steps of the local method.
(d) The processing unit of the local method.
The biggest difference between local and global is that in the local unit, the neural signal $\textbf S^{(1)}_{1:k_1, left}$ corresponding to $\overline x^{(1)}_{1:k_1, left}$ is processed by an unsupervised algorithm (Un-EM) to obtain $\overline x'^{(1)}_{1:k_1, left}$, instead of the source $\overline x^{(1)}_{1:k_1, left}$.}
    \label{fig:ViF-SD2E_gl}
    \vspace{-10pt}
\end{figure}

Here, we use the term `interaction learning' to refer to a learning method realized by interaction between the brain and the external world.
That is, we refer to the ViF-SD2E's learning as `interactive learning with weak-supervision.'
ViF-SD2E includes local and global methods in its open-loop and closed-loop modes.
For the SD module in Figure \ref{fig:ViF-SD2E_diagram}, we can use the global method or the local method as introduced in Figure \ref{fig:ViF-SD2E_gl}. 
Both the global and the local methods are two independent alternatives. 

\subsubsection{The exploration from 2E strategy}

The purpose of the exploration is to mine as much as possible the hidden paradigm in neural signals.
The unsupervised exploration has an iterative process to convergence.
When the exploration's parameters are updated via the exploitation's output, ViF-SD2E is in closed-loop; when the exploration's parameters are updated via self-iteration, ViF-SD2E is in open-loop.
In this paper, an EM algorithm is used, which is based on the SSM principle.
In the iterative process, the E-step and M-step are defined \cite{smith2005erratum}.
The weight is updated as \cite{feng2020weakly}:
\begin{eqnarray}
\ &\bf\tilde{W} &= {\left[\frac{K\sum_{k=1}^K{\bf{S}}_k\hat z_k - \sum_{k=1}^K\hat z_k\sum_{k=1}^K{\bf{S}}_k}{\sum_{k=1}^K\left(\hat z_k^2 + P_k\right) - \sum_{k=1}^K\hat z_k}, \frac{1}{K}\left(\sum\nolimits_{k=1}^K{\bf{S}}_k - {\bf\bar a}\sum\nolimits_{k=1}^K\hat z_k\right)\right]}^\mathrm{T}
\end{eqnarray}
where $\bf\tilde{W}$ is the updated weight.
$\bf\bar{a}$ is equivalent to $\bf{a}$ in $\bf\tilde{W} = [\bf{a},\bullet]$.
$\hat z_k$ is the position output by the exploitation in each iteration of ViF-SD2E (closed-loop) or the predicted position in each interation of the EM (open-loop).
${\bf{S}}_k$ are the input neural signals.
$K$ is the data length.
$P_k$ is the covariance of $\hat z_k$ at time $k$.

\subsubsection{The global method and local method from robust SD module}

Figure \ref{fig:ViF-SD2E_gl} shows the processing steps of the global and local methods.
The functions of the global and local methods are obtained by the infinite expansion of the global unit and local unit, respectively.
It can be seen from the figure that after each processing unit, the space where the predicted/corrected values are located is divided into two parts, left and right in the $x$-axis or up and down in the $y$-axis.
When $N=0$, ViF-SD2E is an unsupervised model without interaction.
As $N$ increases, the corrected values will gradually approach the subspace where the actual labels are located on the $x$-axis or the $y$-axis.
In the global method, neural signals are not reused for movement prediction in the SD module.
In the local method, neural signals complete their self-organization via 0/1 signals for movement prediction of the next subspace.
Our goal is to keep the corrected values and the actual labels in one subspace from $N=1$ to $N=+\infty$.
So, the main difference between the local method and the global method is whether the neural signals are reused or not in each subspace.
Finally, the output corrections in each subspace are restored to the original space.

\subsubsection{The exploitation from 2E strategy}

The purpose of the exploitation is to memorize or store the information (knowledge); that is, the exploration uses the corrected values and the original input neural signals to train an algorithm.
In the training, the output of the exploitation is used to update weight $\bf\tilde{W}$ of the exploration (closed-loop) or else not used (open-loop).
In the testing, the trained exploitation is used for testing and can also prove the $N$-confidence of the corrected position.
Here, we choose a classic LSTM as the network skeleton to prove the effectiveness of our work.

Finally, we believe that the main point of the proposed ViF-SD2E should be the robustness and fault tolerance shown by the ViF-SD2E so that the trained exploitation can reach the level of accuracy of a supervised one.
Also, the robust sequence 0/1 signals from the ViF enter the SD2E to make the neural signals of the brain have the ability to self-organize themselves for movement prediction (the local method).

\section{Experiments}
\label{experiment}
\subsection{Data set collection}
\label{data}

The data was collected and can be found at https://booksite.elsevier.com/9780123838360/ $\rightarrow$ Chapter Materials $\rightarrow$ Chapter 22 $\rightarrow$ Chap22\_ContinuousTrain (data 1) and Chap22\_ContinuousTest (data 2) \cite{wallisch2014matlab}.
The sampling period is 70 ms.
This is about 14.286 Hz, and the total sampling time of 3103 samples is about 217.21 seconds.
The specific characteristics are as follows: \textbf{Data 1:} The feature matrix is size $K_1\times42$ in data 1; it has $K_1$ rows and 42 columns, where $K_1=3103$ is the number of samples, and the feature dimension (number of neural signals) of each sample is 42.
The position matrix is of size $K_1\times2$, it has $K_1$ rows and 2 columns, where the first column is the $x$-axis, and the second column is the $y$-axis.
\textbf{Data 2:} The format of data 2 is similar to that of data 1.
The feature matrix is size $K_2\times42$.
$K_2=3103$ is the number of samples, and each sample's feature dimension (number of neural signals) is 42.
The position matrix is of size $K_2\times2$.
The first column is the $x$-axis, and the second column is the $y$-axis.
For other introductory observations, refer to \cite{feng2020weakly}.
Each sample has two labels containing the $x$- and $y$-position.
Because the labels of the $x$- and $y$-axis are independently predicted, we do the following: \textbf{Experiment A:} Data 1 (3103 $x$-axis samples + 3103 $y$-axis samples) is used as the training set, and Data 2 (3103 $x$-axis samples + 3103 $y$-axis samples) is used as the testing set.
\textbf{Experiment B:} Data 2 is used as the training set, and Data 1 is used as the testing set.
The evaluation measure is the root mean square error.

\subsection{Decoding results and analysis}
\label{decoding}

\begin{figure}   
    \centering
    \subfigure[$x$-axis]{
        \includegraphics[width=0.29\textwidth]{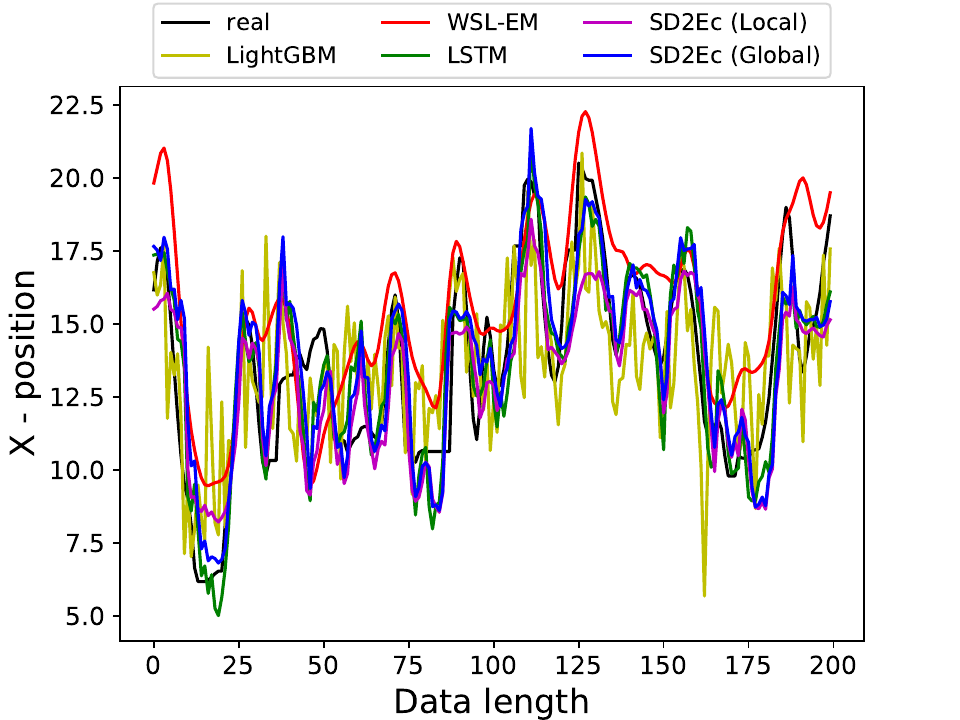}
    }
    \subfigure[$y$-axis]{
        \includegraphics[width=0.29\textwidth]{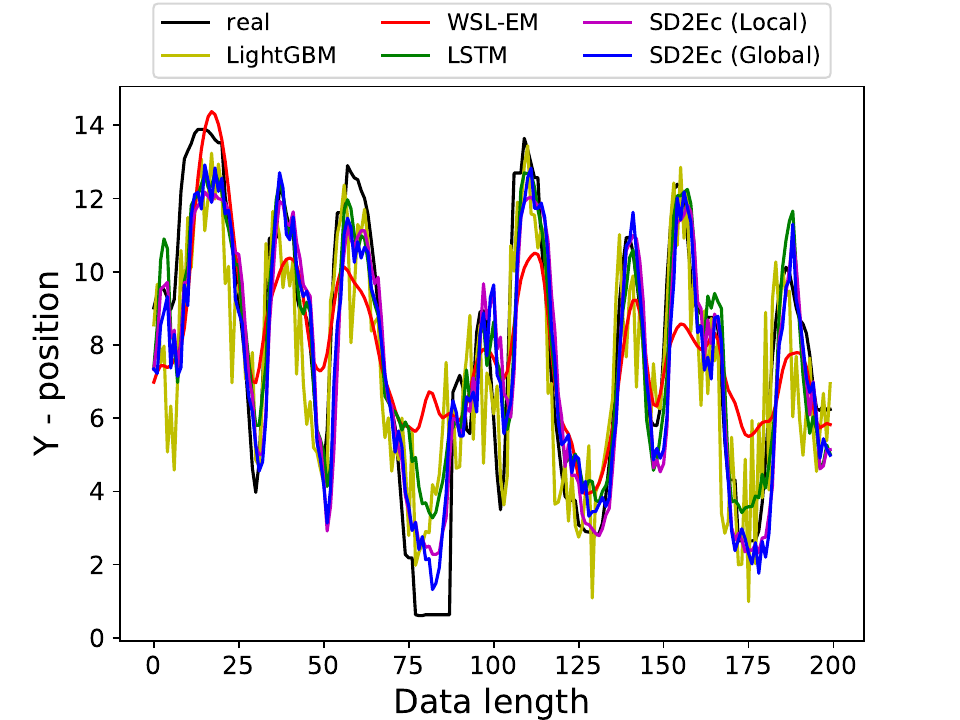}
    }
     \subfigure[$xy$-plane]{
        \includegraphics[width=0.29\textwidth]{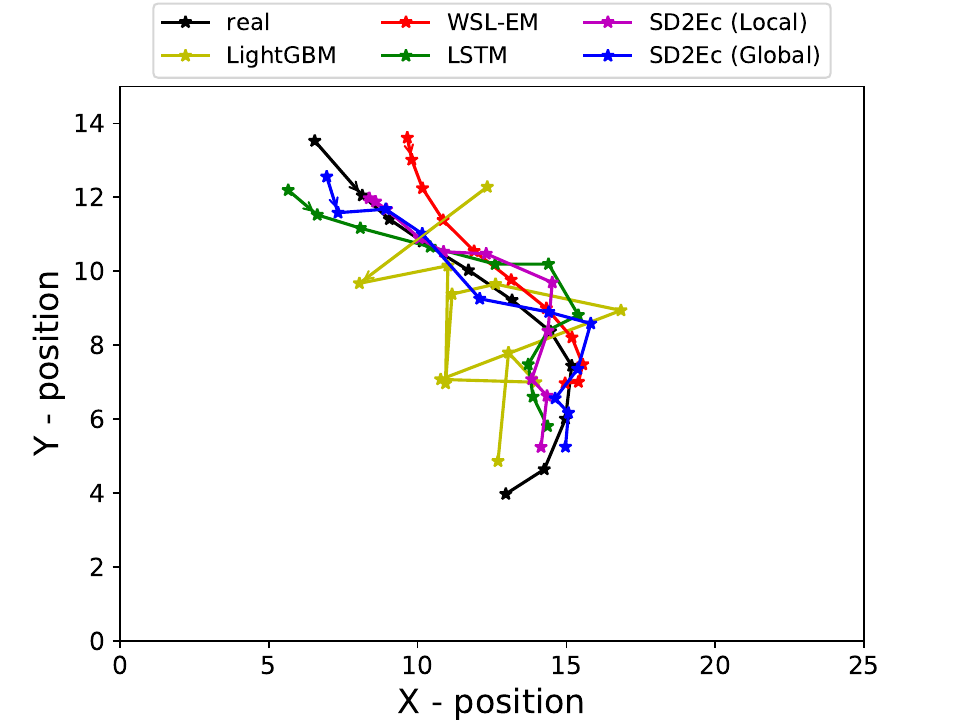}
    }
    \caption{The movement trajectory decoded via neural signals in Experiment A.}
    \label{fig:ViF-SD2E_movement}
\end{figure}

\newcommand{\tabincell}[2]{\begin{tabular}{@{}#1@{}}#2\end{tabular}}
\begin{table}
  \caption{The comparison experiment for ViF-SD2E and other algorithms in Experiments A and B}
  \label{sampletable}
  \centering
  \begin{tabular}{lcccccccc}
    \toprule
    Algorithm & \tabincell{c}{Experiment\\$(x, y)$A} & \tabincell{c}{RMSE\\$(xy)$A} & \tabincell{c}{Experiment\\$(x, y)$B} & \tabincell{c}{RMSE\\$(xy)$B} & \tabincell{c}{RMSE $(xy)$\\(A+B)/2}\\
    \midrule
    KF \cite{wallisch2014matlab}   & (4.569, 2.974) & 5.452 & (3.602, \textbf{1.649})	& 3.962 & 4.707 \\
    XGBoost \cite{chen2016xgbost}	& (3.789, 2.590) & 4.590 & (3.753, 2.103) & 4.302 & 4.446 \\
    LightGBM \cite{ke2017ligbm} & (3.841, 2.464) & 4.563 & (3.735, 2.073) & 4.272 & 4.178 \\
    WSL-KF \cite{feng2020weakly} & (5.403, 3.202) & 6.281 & (3.703, 1.754)	& 4.097 & 5.189 \\
    WSL-EM \cite{feng2020weakly} & (4.518, 2.707) & 5.267 & (3.398, 2.074)	& 3.981 & 4.624 \\
    LSTM \cite{pan2018rapid} & (3.384, 2.330) & 4.109 & (\textbf{3.337}, 1.829) & \textbf{3.805} & 3.957 \\
    \midrule \midrule
    \tabincell{l}{ViF-SD2Ec (Local)} & (3.341, 2.296) & 4.054 & (3.448, 1.916) & 3.945 & 4.000 \\
    \tabincell{l}{ViF-SD2Ec (Global)} & (3.362, \textbf{2.191}) & \textbf{4.013} & (3.386, 1.900) & 3.883 & \textbf{3.948} \\
    \tabincell{l}{ViF-SD2Eo (Local)} & (\textbf{3.312}, 2.302) & 4.033 & (3.436, 1.915) & 3.934 & 3.984 \\
    \tabincell{l}{ViF-SD2Eo (Global)} & (3.394, 2.215) & 4.053 & (3.504, 1.902) & 3.987 & 4.020 \\
    \bottomrule
  \end{tabular}
  \label{fig:ViF-SD2E_comparison}
\vspace{-10pt}
\end{table}

Figure \ref{fig:ViF-SD2E_movement} shows the movement of the $x$- and $y$-axes for the 200 samples selected in Experiment A.
Firstly, in Figure \ref{fig:ViF-SD2E_movement}(a), the decoded movement of the ViF-SD2Ec almost well follow the real trajectory, like that of supervised LSTM on the $x$-axis.
Next, we will focus on the weakly-supervised methods.
Although WSL-EM tracks the movement, it does not do this as well as the other algorithms, such as the 10-30, 60-80, and so on.
Secondly, in Figure \ref{fig:ViF-SD2E_movement}(b), the movement on the $y$-axis is given.
The performance of each algorithm on the $y$-axis is almost the same as that of each algorithm on the $x$-axis.
Finally, in Figure \ref{fig:ViF-SD2E_movement}(c), 11 samples are selected from Figure \ref{fig:ViF-SD2E_movement}(a) and \ref{fig:ViF-SD2E_movement}(b) for visualization in the $xy$-plane.
Overall, the decoded movement in the ViF-SD2Ec seems to be a little closer to the real trajectory.

Table \ref{fig:ViF-SD2E_comparison} lists the errors of the ViF-SD2E (ViF-SD2Ec and ViF-SD2Eo) compared with other algorithms in A and B.
Firstly, the minimum error of the ViF-SD2Ec (Global) is 3.948 in the mean error of (A+B) and is about 0.2\% lower than that of the supervised LSTM, which implies that the trained exploitation (LSTM) has also reached a supervised level, and further indicates that the corrected position has higher $N$-confidence.
Then, ViF-SD2E in Experiment A shows better performance.
However, ViF-SD2E in Experiment B does not improve the performance well.
A possible reason for this is that the training samples in Experiment B are more diverse for supervised mode, or perhaps ViF-SD2E's parameters have not been adjusted to the optimal level
(see subsection \ref{spatial parameter}).
Finally, there are no large differences between ViF-SD2Eo and ViF-SD2Ec in decoding the movement with the local and global methods, but the calculational burden of ViF-SD2Eo is higher than that of ViF-SD2Ec
(see subsection \ref{computational complexity}).

\subsection{Spatial parameter and ablation experiment in ViF-SD2E}
\label{spatial parameter}

\begin{figure}    
    \centering
    \subfigure[The error curve of ViF-SD2Ec in A.]{
        \includegraphics[width=0.32\textwidth]{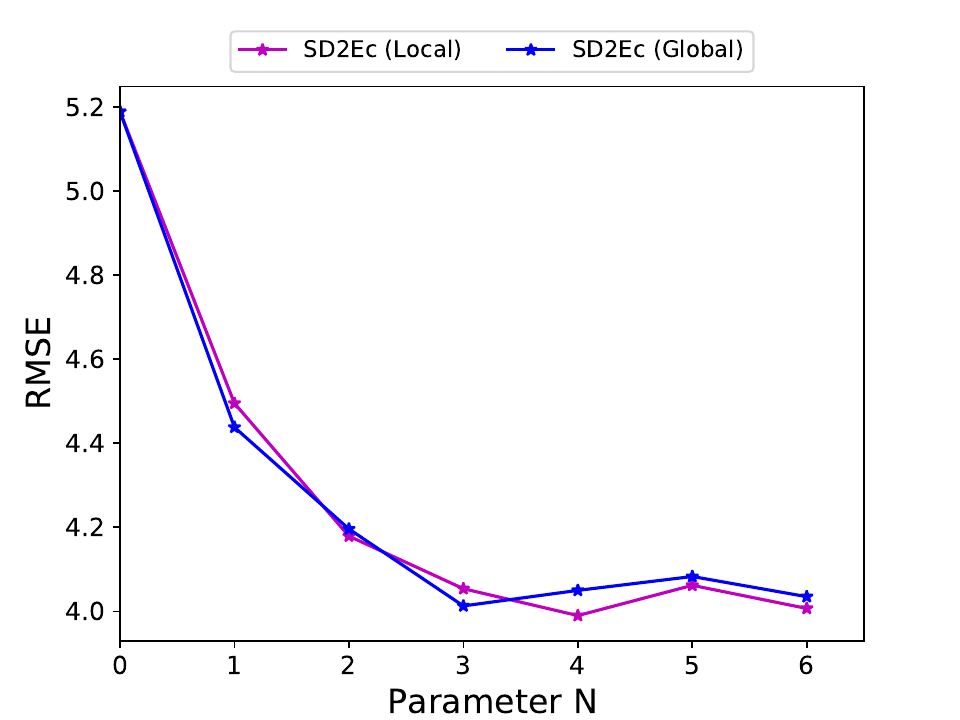}
    }
    \subfigure[The error curve of ViF-SD2Ec in B.]{
        \includegraphics[width=0.32\textwidth]{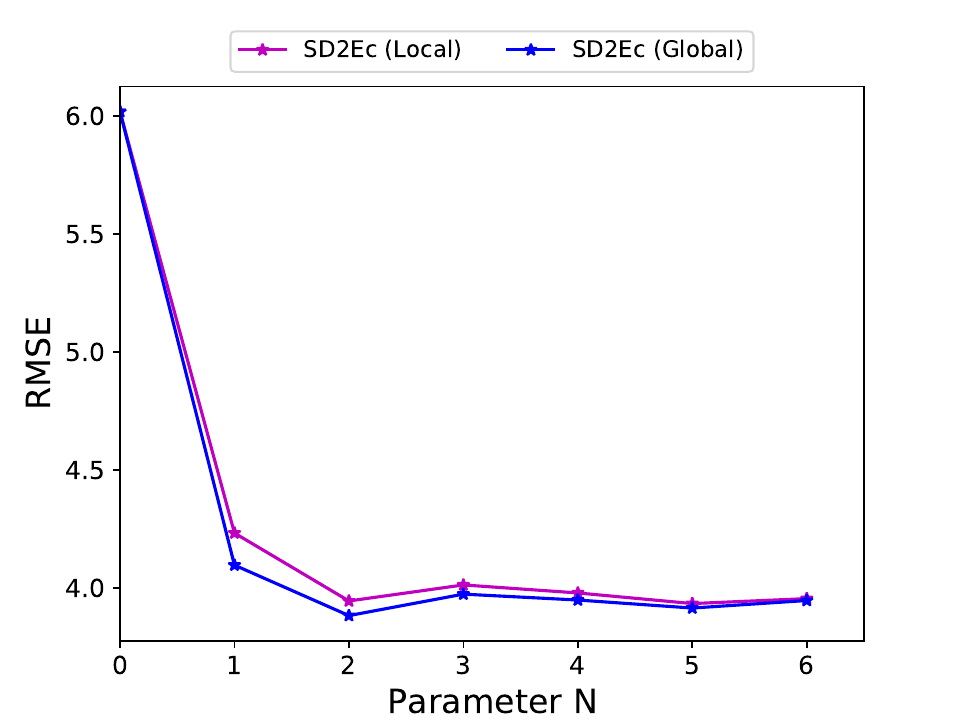}
    }\\
    \centering
    \subfigure[The error curve of ViF-SD2Eo in A.]{
        \includegraphics[width=0.32\textwidth]{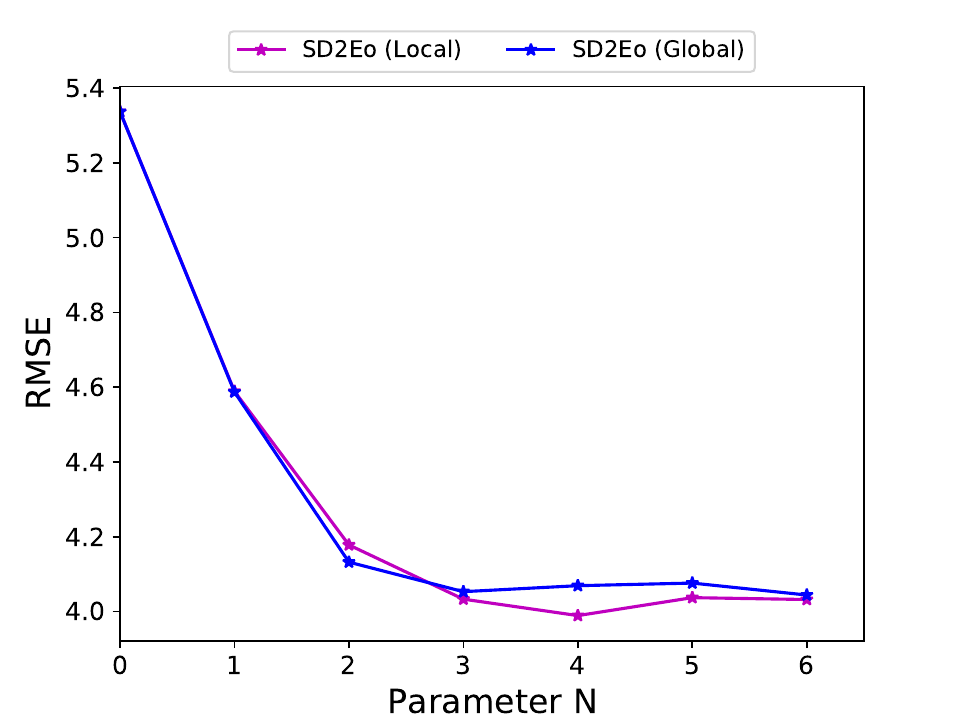}
    }
    \subfigure[The error curve of ViF-SD2Eo in B.]{
        \includegraphics[width=0.32\textwidth]{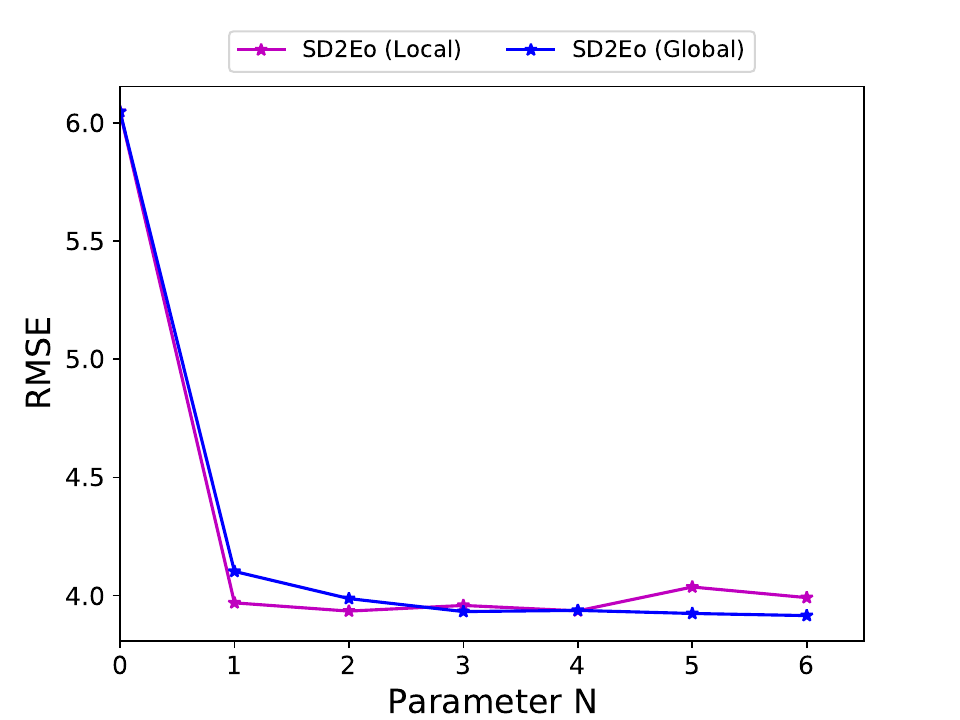}
    }
    \caption{The error curve of the ViF-SD2Ec and ViF-SD2Eo with parameter $N$ in Experiments A and B.}
    \label{fig:ViF-SD2E_parameterAN}
\end{figure}

\begin{wraptable}{tr}{0.50\textwidth} 
\vspace{-18pt}
    \caption{Ablation for ViF-SD2Ec in Experiment A} 
	\centering
	\begin{tabular}{lcc}
    	\toprule
        Algorithm & \tabincell{c}{RMSE $(x, y)$} & \tabincell{c}{RMSE $(xy)$} \\
        \midrule
        Un-EM & (6.870, 3.610) & 7.761 \\
        Un-EM \& LSTM & (4.403, 2.949) &  5.298\\
        Un-EM \& SD (Local) & (4.066, 2.907) & 4.998 \\
        Un-EM \& SD (Global) & (4.208, 2.771) & 5.038 \\
        \tabincell{l}{ViF-SD2Ec (Local)} & (\textbf{3.341}, 2.296) & 4.054  \\
        \tabincell{l}{ViF-SD2Ec (Global)} & (3.362, \textbf{2.191}) & \textbf{4.013}  \\
        \bottomrule
    \end{tabular}
    \label{fig:ViF-SD2E_ablation}
\vspace{-6pt}
\end{wraptable}

Figure \ref{fig:ViF-SD2E_parameterAN} shows the error curves of ViF-SD2E (ViF-SD2Ec and ViF-SD2Eo) in Experiments A and B when $N$ increases.
As shown in Figure \ref{fig:ViF-SD2E_parameterAN} (a) to (d), when $N=1$, the decoding errors have dropped sharply because the unsupervised mode with $N=0$ has changed to the weakly-supervised mode with $N=1$.
After $N=3$ or $N=4$, the error basically fluctuates at a low level until convergence.
From the above, since the decoding errors may decline if we continue to increase $N$, the ViF-SD2E may outperform or be close to the supervised LSTM.
In addition, the initial decline of ViF-SD2Eo seems to be faster than that of ViF-SD2Ec from $N=0$ to $N=3$.
Finally, it should be noted that $N=3$ is an appropriate convergence and is what has been set throughout this paper.
That is, $N=3$ has higher confidence.

Table \ref{fig:ViF-SD2E_ablation} lists the ablation experiment error for weakly-supervised ViF-SD2Ec in Experiment A.
The exploration (Un-EM) is indispensable to explore the hidden paradigm of the neural signal.
Therefore, Un-EM and Un-EM\&LSTM are unsupervised methods; Un-EM\&SD and ViF-SD2Ec are weakly-supervised methods.
From the results of the ablation experiment, the decoding error of ViF-SD2Ec (Global) is about 48.3\% lower than that of the unsupervised Un-EM algorithm, is about 24.3\% lower than that of the unsupervised Un-EM\&LSTM, is about 19.7\% lower than that of the weakly-supervised Un-EM\&SD (Local), and is about 20.3\% lower than that of the weakly-supervised Un-EM\&SD (Global).
This implies that robust ViF-SD2Ec has certain advantages in neural decoding.

\subsection{Corrected values and robustness in ViF-SD2E}
\label{Correction and robustness}

\begin{table}
  \caption{The correction experiment for training and testing in Experiments A and B}
  \label{sample-table}
  \centering
  \begin{tabular}{lcccccccc}
    \toprule
    Algorithm & \tabincell{c}{Uncorr/corr\\train($xy$) A} & \tabincell{c}{RMSE\\Test($xy$) A} & \tabincell{c}{Uncorr/corr\\train($xy$) B} & \tabincell{c}{RMSE\\Test($xy$) B} & \tabincell{c}{RMSE($xy$)\\(A+B)/2\\Uncorr/corr/Test}\\
    \midrule
    \tabincell{l}{ViF-SD2Ec (L)} & 6.585/1.491 & 4.054 & 10.236/1.171 & 3.945 & 8.411/1.331/4.000 \\
    \tabincell{l}{ViF-SD2Ec (G)} & 6.585/1.463 & 4.013 & 10.236/1.455 & 3.883 & 8.411/1.459/3.948 \\
    \tabincell{l}{ViF-SD2Eo (L)} & 7.141/1.491 & 4.033 & 10.366/1.171 & 3.934 & 8.754/1.331/3.984 \\
    \tabincell{l}{ViF-SD2Eo (G)} & 7.141/1.506 & 4.053 & 10.366/1.378 & 3.987 & 8.754/1.442/4.020 \\
    \bottomrule
  \end{tabular}
  \label{fig:ViF-SD2E_correction}
\end{table}

\begin{figure}[t]
\makeatletter\def\@captype{figure}\makeatother
\begin{minipage}[hbp]{0.47\textwidth}
    \centering
    \includegraphics[width=0.56\textwidth]{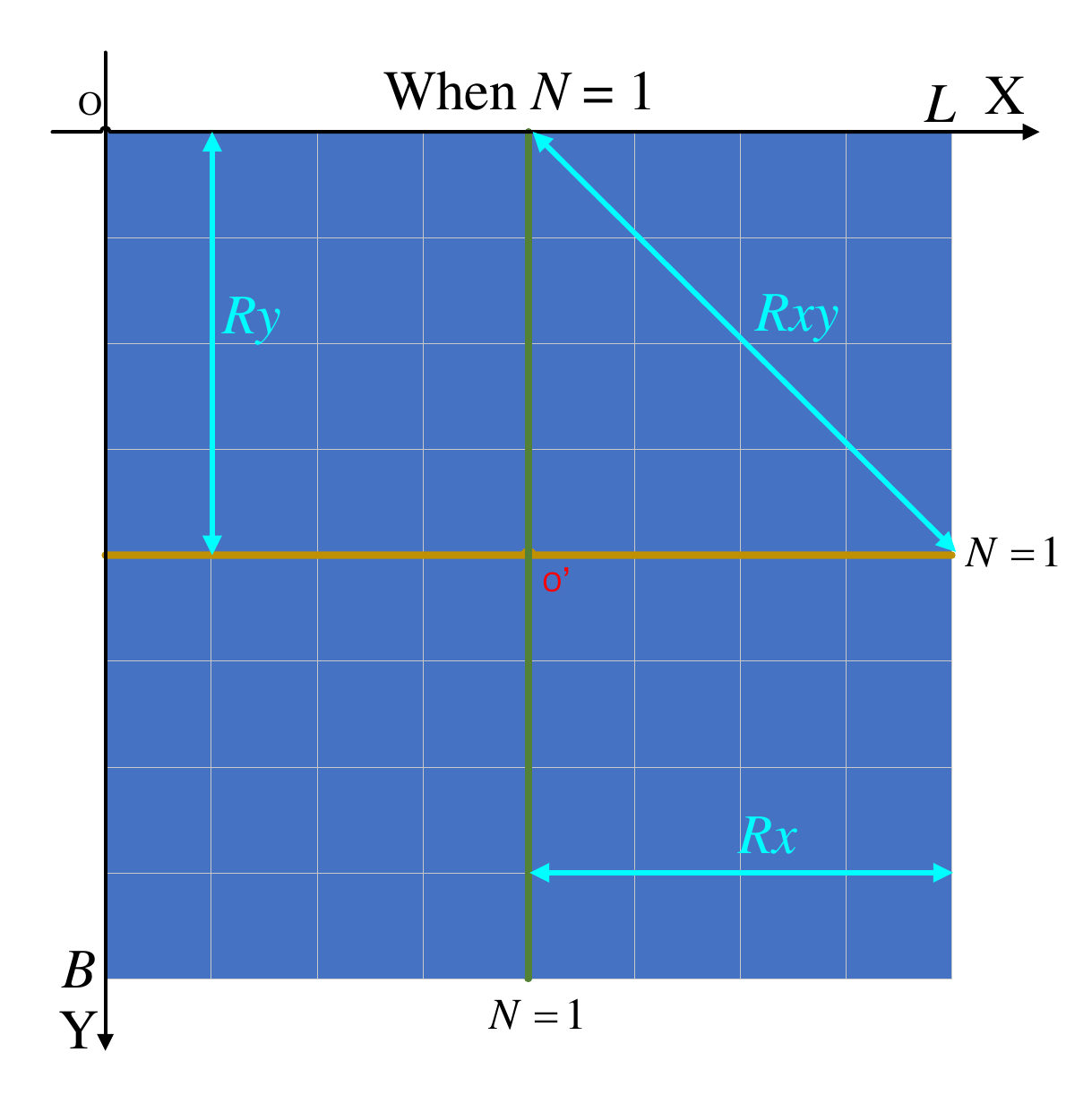}
    \caption{The fault tolerance $R$ in ViF-SD2E.}
    \label{fig:ViF-SD2E_robust}
\end{minipage}
\makeatletter\def\@captype{table}\makeatother
\begin{minipage}{0.54\textwidth}
    \centering
    \caption{Robustness in ViF-SD2E ($L\approx 25$, $B\approx 15$)} 
    \begin{tabular}{lccccc}
	\toprule
    Parameter & \tabincell{c}{$R_x$ ($L/2^N$)} & \tabincell{c}{$R_y$ ($B/2^N$)} & \tabincell{c}{$R_{xy}$} \\
    \midrule
    $N=0$ & 25 & 15 & 29.155 \\
    $N=1$ & 12.5 & 7.5 & 14.577 \\
    $N=2$ & 6.25 & 3.75 & 7.289 \\
    $N=3$ & 3.125 & 1.875 & 3.644 \\
    $N=4$ & 1.563 & 0.938 & 1.823 \\
    $N=5$ & 0.781 & 0.469 & 0.911 \\
    $N=6$ & 0.391 & 0.234  & 0.456 \\
    \bottomrule
    \end{tabular}
    \label{fig:ViF-SD2E_robustN}
\end{minipage}
\vspace{-3pt}
\end{figure}

Table \ref{fig:ViF-SD2E_correction} gives the uncorrected and corrected errors in the training and testing errors in the test.
From Table \ref{fig:ViF-SD2E_correction}, that after being corrected, the errors are concentrated at about 1.5.
In the (A+B)/2, the test error is about half of the uncorrected error and about three times the correction.
Therefore, compared with traditional supervision, ViF-SD2E may make some experimental errors in training. 
But such small training errors seem to have little effect on the generalization ability (the supervised level can still be reached in testing) in the  Table \ref{fig:ViF-SD2E_comparison} and Table \ref{fig:ViF-SD2E_correction}.
In addition, Figure \ref{fig:ViF-SD2E_parameterAN} shows that when $N=1$, the decoding error will drop sharply; when $N=3$, the decoding error tends to be stable generally.
These also imply that ViF-SD2E has strong robustness. 
Refer to the following analysis for maximum fault tolerance, and Figure \ref{fig:ViF-SD2E_movementspace-2} for spatial encoding.

In Figure \ref{fig:ViF-SD2E_robust}, $R$ is an evaluation value, which is also called maximum fault tolerance.
Here, $R_x$, $R_y$ and $R_{xy}$ mainly depend on the sizes of the active space (length $L$ and width $B$). 
For prediction of the $x$-coordinate, the length $L$ will be equally divided into two segments ($N=1$), with each segment encoded as 0 and 1. 
Thus, the maximum fault tolerance of each segment is $R_x=L/2$. 
Next, we make predictions for each segment, and continue to divide them equally ($N=2$). 
Now, the maximum fault tolerance of each segment becomes $R_x=L/4$. 
Going forward continuously, $R_x=L/{2^N}$. Similarly, the maximum fault tolerance of $y$-coordinate is $R_y=B/{2^N}$. 
Finally, they are restored to a $xy$-plane, $R_{xy}=(R_x^{2}+R_y^{2})^{1/2}$ (Euclidean distance).
Table \ref{fig:ViF-SD2E_robustN} shows the maximum fault tolerance $R$ from $N=0$ to $N=6$.
The larger the $N$, the smaller the $R$, meaning that the robustness of the model becomes weak (therefore, the accuracy of the label is required to be further raised). 


\subsection{Complexity (running time) and limitation of ViF-SD2E}
\label{computational complexity}

The complexity of ViF-SD2E depends on that of the exploration and exploitation and the SD module.
Of these, the SD module contains the accumulation of the multiple exploration (or an independent algorithm) and the number of divisions of the space into subspaces.
Therefore, it is important to reduce the complexity of the algorithms used.
In addition, it also seems to be a P/NP problem in terms of exponentially increasing space size and running time.
However, the most ideal state is that the purpose of the ViF-SD2E is to obtain the best/optimal values (or something close to the actual labels) when $N$ is the smallest.
In this paper, we assume that the exploration's iterations are $n_1$, and each iteration time is $t_1$; the exploration's iterations are $n_2$, and each iteration time is $t_2$; SD is divided into $n_3$ subspaces, and each run time is $t_3$.
Then, the consumption of ViF-SD2E (closed-loop) is about $(t_1 + {t_2}n_2 + {t_3}n_3)n_1$; the consumption of ViF-SD2E (open-loop) is about ${t_1}n_1 + {t_2}n_2 + {t_3}n_3$.
Therefore, ViF-SD2Ec consumes much more time than ViF-SD2Eo, which is about ${t_2}n_1(n_2 - 1) + {t_3}n_1(n_3-1)$.

In addition, in ViF-SD2E, the exploration and exploitation can adopt any unsupervised algorithm and supervised algorithm respectively in this paper, and the value of $N$ is initially set. We can ensure that ViF-SD2E is convergent as $N$ increases, see Figure \ref{fig:ViF-SD2E_parameterAN}.
In order to maximize the performance of the model one could develop a new algorithm that includes these ideas; for example, making $N$ adaptive. 
In our opinion, this can be a larger workload due to the immaturity of the technology, as it seems to involve carrying out Bayesian inference using Exploitation (the prior turns into a posterior, which turns into a prior again), see Figure \ref{fig:ViF-SD2E_Procedure}. Moreover, extrapolating from the current results, the Bayesian pattern of the model is highly likely to be correct.

Based on the above analysis, we give the advantages and disadvantages of open-loop and closed-loop.
On the whole, both open-loop and closed-loop methods can deliver a complete algorithm design on their own.
Open-loop has the advantage of higher operating efficiency. 
On the downside, however, exploration does not depend on experienced exploitation when updating parameters; in the case of closed-loop, exploration relies on empirical exploitation, which may provide finer prediction accuracy, but with a disadvantage of high computational complexity. 
However, we have not yet found a basis to decide which of the open-loop and closed-loop methods is more reasonable in neuroscience. They are a limitation of VIF-SD2E and are still expected to be further validated by computational neuroscience and neurophysiology. 

\section{Discussion}
\subsection{Inspiration for the designed ViF-SD2E}
The core origin of the designed ViF-SD2E in this work stems from a factual observation that unsupervised decoded motion trajectories and true trajectories illustrate a symmetric phenomenon in the activity space.
Given that datasets are derived from brain neural signals, it is believed that such an observation may have certain key information hidden.
Based on this observation, a more intuitive ViF-SD2E is designed to simulate the neural decoding process of the brain, and then the implicit information of the model is mined and interpreted.
Last but not least, the main feature of the ViF-SD2E is that it can make interaction via encoded 0/1 signals and then correct the system's own predictions, seemingly implying the existence of cognitive patterns of fuzzy interaction in the nervous system. 
As a result, ViF-SD2E may be of great value for the development of fuzzy systems, which mimic human brain's functions.

\subsection{Broader impacts in weak-labels and regression}
When are the weak labels needed? 
Here, the encoded 0/1 signals can help us focus on which area is the prediction/label in a fixed space or subspace, and the $N$-value has the effect of adjusting the self-attention (the resolution of the location) via 0/1 signals. For example, compared with human behavior, people seem to have strong robustness and many ambiguities in learning, just as we can estimate where black dots may appear on a computer screen, but we cannot give specific coordinates. In our opinion, this mode is in line with the coding characteristics of 0/1 in ViF-SD2E.
Can regression and classification be linked? 
All regression labels are converted to 0/1 signals by ViF and then fed into the SD2E. 
In addition, the SD2E's output will also be converted into 0/1 signals to interact or compare with the given 0/1 signals from ViF, and 0/1 is a classification idea.
Beside, we believe it may also have a key relationship with neural coding or the distribution of the data.
If the symmetry between the decoded unsupervised values and the actual labels is not perfect, the $N$-value can be increased to reduce errors/robustness.

\section{Conclusion}
\label{conclusion}

In this paper, we proposed a robust weakly-supervised method, called ViF-SD2E, for neural decoding.
An encoded sequence of 0/1 signals replaces the actual labels in the movement space/subspace and are used as feedback between ViF and SD2E.
The potential significance of this paper are: 1) we designed a universal ViF and SD module to enhance the application of the spatial symmetry between the movement predictions and the actual labels; 2) an exploration is employed to explore the original neural signals, and an exploitation is used to make effective use of the corrected values for future calls to evaluate $N$-confidence and is used for prediction, which is referred to as exploration--exploitation (2E); 3) we adopt an unsupervised EM and a supervised LSTM with time-series to decode the movement.

We conducted extensive experiments to verify the effectiveness of our method. 
Beside, the degree of supervision can be flexibly controlled via a parameter $N$. When $N=0$, the model degenerates into an unsupervised mode, but is equivalent to a supervised model when $N\rightarrow + \infty$. So, choosing an appropriate $N$ can enable the model to fully use the 0/1 signals robustly for noisy target values or movements with recorded partial changes in the measurement setups. Finally, if ViF-SD2E together with only neural signals involved in the calculation is regarded as a human brain, we believe that the ultimate goal of ViF-SD2E is to obtain the optimal/satisfactory prediction (or something close to the actual label) when $N$ is the smallest; just like a baby's learning process or our learning process when faced with new things.

In the future, we will continue to explore the Bayesian pattern of ViF-SD2E in addition to the cognitive pattern of fuzzy interaction of this model.


\medskip
{
\small

\bibliographystyle{unsrt}

}



\appendix

\section{Parameter and running environment}
\label{parameter}
The operating environment of KF, XGBoost, LightGBM, WSL-KF, and WSL-EM is as follows. The PC operation system running the decoding algorithms is Windows 10 professional 64-bit DirectX 12, the processor is Intel Corei7-9700@3.00GHz eight-core, and the processing software is MatLabR2020b\_Win. Also, the operating environment of the LSTM and ViF-SD2E is as follows. The 4U rack server operating system running the decoding algorithms is UBUNTU 20.04.1, the barebone system is SYS-4029GP-TRT, the processor is Intel Xeon Gold 6226R 2.90G/16Core/22M/150W, the GPU computing card is NVIDIA RTX 3090 24G memory, and the processing software is python 3.8.8 and MatLab R2020b\_Linux. Moreover, the parameters were set as follows.

\textbf{KF:} The initial position $z_0$ = 0, covariance $P_{y,0\mid0}$= 10, state noise variance $Q_w$= 0.8, and observed noise variance ${\bf R}_v={\mathbf{(\bf\bar{s}-\bf\hat{s})}^\mathrm{T}(\bf\bar{s}-\bf\hat{s})}/{(K-1)}$, where ${\bf\bar{s}}=\mathbf{({\bf s}_0,{\bf s}_1,..., {\bf s}_{\emph K-1})}^\mathrm{T}$ and $\bf\hat{s}=\bf{Y^\dag\hat{W}}$. \textbf{XGBoost:} The initial $max\_depth$ = 5, $learning\_rate$ = 0.1, $n\_estimators$ = 150, $objective$ = $\rm{'reg:linear'}$, $n\_jobs$ = -1, $eval\_metric$ = $\rm{'logloss'}$, and $verbose$ = 100. \textbf{LightGBM:} The initial $num\_leaves$ = 33, $learning\_rate$ = 0.05, $n\_estimators$ = 150, $eval\_metric$ = $\rm{'logloss'}$, and $verbose$ = 100. \textbf{WSL-KF:} This is the KF in WSL. The initial position $z_0$= 10, covariance $P_{y,0\mid0}$= 10, and system noise variance $R_w$ = 0.8. The observed noise variance $\bf{R}_v$= diag(ones(1,42)) represents generation behavior 1, and the diagonal matrix is listed as 42. The initial weight $\bf\tilde{W}_0$ = ones (2, 42) represents generation behavior 2, and the all-ones matrix is listed as 42. \textbf{WSL-EM:} This is the EM in WSL. The initial position $z_0$= 10, covariance $P_{y,0\mid0}$ = 10, and system noise variance $R_w$ = 2. The observed noise variance $\bf{R}_v$ = diag(ones(1, 42$T$)) represents generation behavior 1 and the diagonal matrix is listed as 42$T$. $\bf\tilde{W}_0$ = ones (2, 42$T$) represents generation behavior 2, the all-ones matrix is listed as 42$T$. The $T$ is set to 10 in the X- and Y-position, and $iteration$ = 8. \textbf{LSTM:} The initial, $input\_size$ = 42$T$, $hidden\_size$ = 70, $output\_size$ = 1, $num\_layers$ = 3, $learning\_rate$ = 0.02. The maximum-value of $epoch$ is set to 1000,and optimization period $t=10$. The Look\_back $T$ is also set to 10 in the X- and Y-position. \textbf{ViF-SD2E (Local) and (Global):} Both of these are robust weakly-supervised methods. The initial position $z_0$= 10, covariance $P_{y,0\mid0}$ = 10, and system noise variance $R_w$ = 2. The observed noise variance $\bf{R}_v$ = diag(ones(1, 42$T$)) represents generation behavior 1 and the diagonal matrix is listed as 42$T$. $\bf\tilde{W}_0$ = ones (2, 42$T$) represents generation behavior 2, the all-ones matrix is listed as 42$T$. The $input\_size$ = 42$T$, $hidden\_size$ = 70, $output\_size$ = 1, $num\_layers$ = 3, $learning\_rate$ = 0.02. The maximum value of $epoch$ is set to 1000. The Look\_back $T$ is also set to 10 in the X- and Y-position, and the value of $iteration$ is set to 8. Finally, the space parameter $N$ is set to 3 in the X- and Y-position. 

For non-deep methods, the experiments of KF, XGBoost, LightGBM, WSL-KF, and WSL-EM can be directly reproduced. For deep LSTM and ViF-SD2E, the underlying LSTM adopted in the exploitation has the same settings as traditional supervised LSTM for the sake of making a fair comparison in the testing.

\section{The smallest computing unit}

\begin{figure*}[h]
\centering
\includegraphics[width=\textwidth]{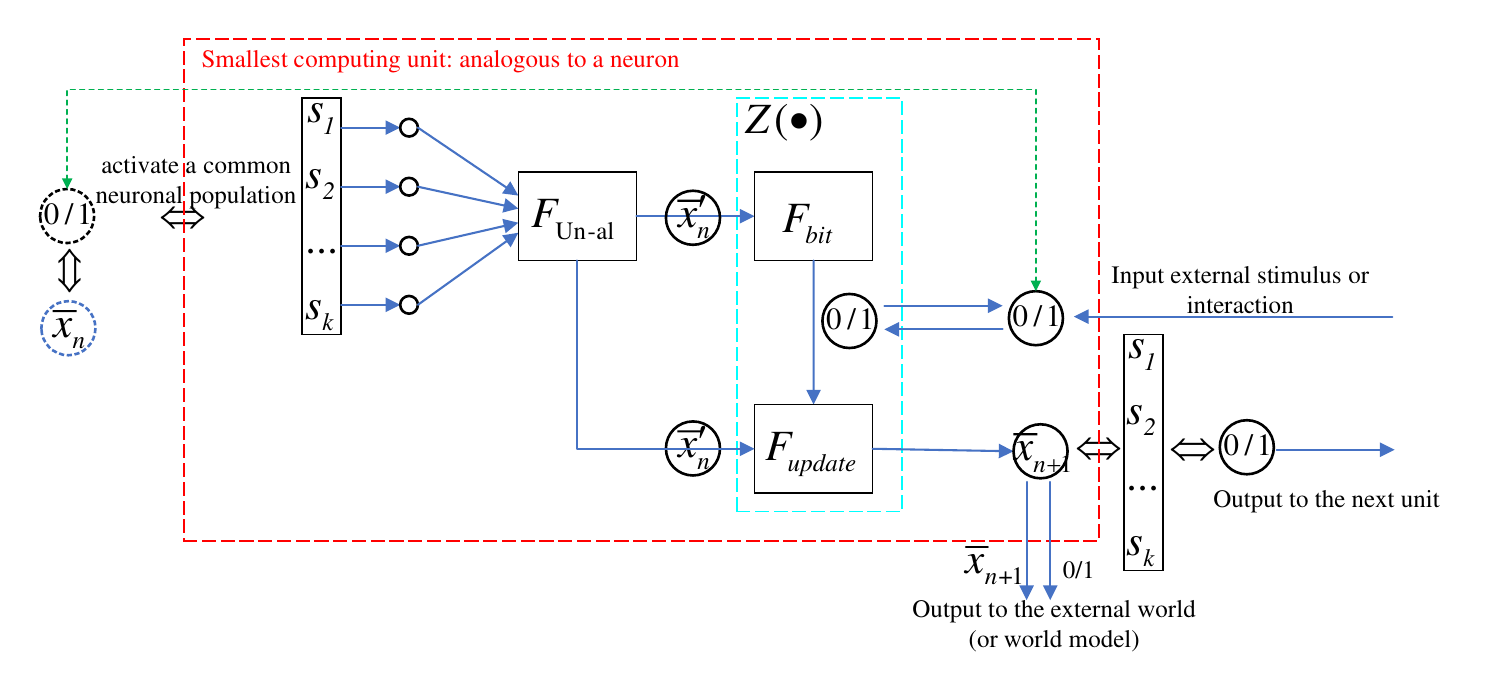}
\caption{The smallest computing unit is from Figure \ref{fig:ViF-SD2E_gl}(c)-(d), which can be compared to a single neuron.
}
\label{fig: Smallest_computing_unit}
\end{figure*}

Finally, to draw an analogy with a single neuron in the brain, the smallest computing unit is showcased in Figure \ref{fig: Smallest_computing_unit}, from Figure \ref{fig:ViF-SD2E_gl}(c)-(d).
This is a computing unit, which can do self-correction function by inputting 0/1 feedback signals.
The computing unit has two outputs: 
the first is the neural signals (or 0/1) corresponding to the corrected prediction value, which is then sent to the next computing unit; and the second is the corrected prediction value or 0/1 symbols, which are used to determine prediction errors based on the exploitation or be outputted directly in the last layer.

\end{document}